\documentclass[11pt]{article} 
\usepackage{rldmsubmit,palatino}
\usepackage{graphicx}
\usepackage{wrapfig}
\usepackage{natbib}
\title{What makes useful auxiliary tasks in reinforcement learning: \\investigating the effect of the target policy}

\author{
Banafsheh Rafiee \\
University of Alberta, Canada, \\
Noah's Ark Lab, Huawei Technologies Canada \\
\texttt{rafiee@ualberta.ca} \\
\And
Jun Jin \\
Noah's Ark Lab, Huawei Technologies Canada\\
\texttt{jun.jin1@huawei.com} \\
\AND
Jun Luo \\
Noah's Ark Lab, Huawei Technologies Canada\\
\texttt{jun.luo1@huawei.com} \\
\And
Adam White\\
University of Alberta, Canada \\
\texttt{amw8@ualberta.ca} \\
}

\begin{document}

\maketitle

\begin{abstract}

Auxiliary tasks have been argued to be useful for representation learning in reinforcement learning. 
Although many auxiliary tasks have been empirically shown to be effective for accelerating learning on the main task, it is not yet clear what makes useful auxiliary tasks. 
Some of the most promising results are on the pixel control, reward prediction, and the next state prediction auxiliary tasks; however, the empirical results are mixed, showing substantial improvements in some cases and marginal improvements in others. 
Careful investigations of how auxiliary tasks help the learning of the main task is necessary.
In this paper, we take a step studying the effect of the target policies on the usefulness of the auxiliary tasks formulated as general value functions. 
General value functions consist of three core elements: 1) policy 2) cumulant 3) continuation function.
Our focus on the role of the target policy of the auxiliary tasks is motivated by the fact that the target policy determines the behavior about which the agent wants to make a prediction and the state-action distribution that the agent is trained on, which further affects the main task learning.
Our study provides insights about questions such as: Does a greedy policy result in bigger improvement gains compared to other policies? Is it best to set the auxiliary task policy to be the same as the main task policy? Does the choice of the target policy have a substantial effect on the achieved performance gain or simple strategies for setting the policy, such as using a uniformly random policy, work as well?
Our empirical results suggest that: 1) Auxiliary tasks with the greedy policy tend to be useful. 2) Most policies, including a uniformly random policy, tend to improve over the baseline. 3) Surprisingly, the main task policy tends to be less useful compared to other policies. 






\end{abstract}


\startmain 

\section{Introduction}
Learning about many aspects of the environment in addition to the main task of maximizing the discounted sum of rewards has been argued to be beneficial in reinforcement learning \citep{sutton2011horde}. 
A common view is that these additional tasks, also known as auxiliary tasks, can improve the data efficiency by shaping the representation \citep{jaderberg2016reinforcement,shelhamer2016loss,mirowski2016learning}.
In environments with sparse reward structures, auxiliary tasks provide instantaneous targets for shaping the representation in the absence of reward. 
It has also been argued that auxiliary tasks can function as regularizers, improving the generalization and avoiding representation overfitting in RL \citep{dabney2020value}. 
Finally, what has been learned about an auxiliary task can be transferred to the main task, improving the data efficiency.

Many auxiliary tasks have been proposed and shown to accelerate learning on the main task. 
However, much of the work on auxiliary tasks have demonstrated their efficacy empirically and in many cases, the empirical results are mixed: in some cases auxiliary tasks result in substantial performance gain over the baselines whereas in other cases they achieve marginal improvements or even harm the performance \citep{jaderberg2016reinforcement, shelhamer2016loss}. 
Recent works have explored how auxiliary tasks benefit the representation learning in a systematic way \citep{lyle2021effect,dabney2020value}. 
More systematic studies are required to answer the question of what makes useful auxiliary tasks. 

In this paper, we take a step toward answer the question of what makes useful auxiliary tasks. 
We specifically consider auxiliary tasks formulated as general value functions (GVF) \citep{sutton2011horde}. 
A core element of GVFs is the target policy as it both determines the target of prediction and the state-action distribution that the agent gets trained on.
We empirically study the effect of the target policies associated with auxiliary tasks on their usefulness. 

\section{Background}
In this paper, we consider the interaction of an agent with its environment at discrete time steps. At each time step $t$, the agent is in a state $S_t \in \mathcal{S}$, performs an action $A_t \in \mathcal{A}$ according to a policy $\pi: \mathcal{A}\times\mathcal{S} \rightarrow [0, 1]$, receives a reward $R_{t+1} \in \mathbf{R}$, and transitions to the next state $S_{t+1}$. 
We consider the problems of prediction and control. 

For the prediction problem, the goal of the agent is to approximate the value function for a given policy $\pi$ where the value function is defined by: $v_{\pi}(s) \doteq \mathbf{E}_\pi[G_t | S_t=s]$ where $G_t = \sum_{k=0}^{\infty}\gamma^k R_{t+k+1}$ is called the return with $\gamma \in [0, 1)$ being the discount factor.
In the control setting, the goal of the agent is to maximize the expected return. 
In this setting, it is common to use state-action value functions: $q_{\pi}(s, a) \doteq \mathbf{E}_\pi[G_t | S_t=s, A_t = a]$.

To estimate the value function, we use semi-gradient temporal-difference learning \citep{sutton1988learning}. 
More specifically, we use TD$(0)$ to learn a parametric approximation $\hat{v}(s; \bf{w})$ by updating a vector of parameters $\bf{w} \in \mathbf{R}^d$ as follows:
$$    \bf{w_{t+1}} \leftarrow \bf{w_t} + \alpha \delta_t \nabla_{\bf{w}} \hat{v}(S_t; \bf{w})
$$

\noindent where $\alpha$ denotes the step-size parameter and $\delta_t$ denotes the TD error: $ R_{t+1} + \gamma \hat{v}(S_{t+1}; \bf{w_t}) - \hat{v}(S_t; \bf{w}_t)$.  $\nabla_{\bf{w}} \hat{v}(S_t; \bf{w})$ is the gradient of the value function with respect to the parameters $\bf{w}_t$. 

For the control setting, to estimate the state-action value functions, we use the control variant of TD$(0)$, Q-learning \citep{watkins1992q}. The update rule for Q-learning is similar to that of TD$(0)$; however, $\hat{q}(S_t, A_t, \bf{w})$ is used instead of $\hat{v}(S_t, \bf{w})$ and the TD error is defined as $ R_{t+1} + \gamma \mbox{max}_a \hat{q}(S_{t+1}, a; \bf{w_t}) - \hat{q}(S_t, A_t; \bf{w}_t)$. 
For action selection, Q-learning is commonly integrated with the epsilon greedy policy.

As the function approximator, we use neural networks. We integrate a replay buffer, a target network, and the RMSProp optimizer with TD$(0)$ and Q-learning as is commonly done to improve performance when using neural networks as the function approximator \citep{mnih2013playing}. 

To formulate auxiliary tasks, a common approach is to use general value functions (GVFs). GVFs are value functions with a generalized notion of target and termination. More specifically, a GVF can be written as the expectation of the discounted sum of any signal of interest: $v_{\pi, \gamma, c}(s) \doteq \mathbf{E}[\sum_{k=0}^{\infty}(\prod_{j=1}^k\gamma(S_{t+j}))c(S_{t+k+1}) | S_t=s, A_{t:\infty} ~ \pi]
$
where $\pi$ is the target policy, $\gamma$ is the continuation function, and $c$ is a signal of interest. General state-action value function $q_{\pi, \gamma, c}(s, a)$ can be defined similarly with the difference that the expectation is conditioned on $A_t = a$ as well as $S_t = s$.

\section{Investigating the effect of the auxiliary task's policy on its usefulness}
\begin{figure}
\centering
  \includegraphics[width=0.6\linewidth]{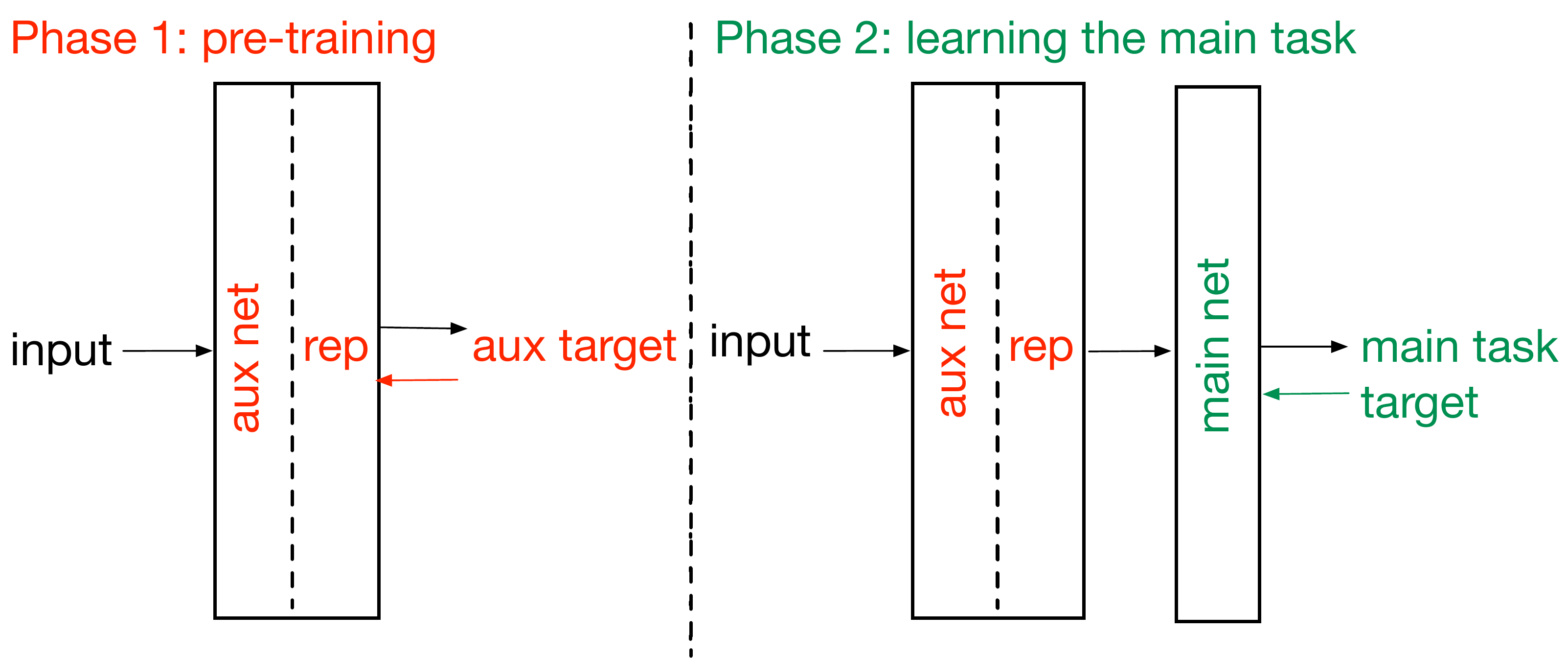}
  \caption{The two phases of pre-training on the auxiliary tasks and learning the main task.}
  \label{fig:phases}
\end{figure}

As we discussed in Section 1, we study the effect of the target policy of the auxiliary tasks on their usefulness. Our experimental setup includes two phases: 1) Pre-training on the auxiliary tasks 2) Learning the main task. For the pre-training phase, a neural network is used as the function approximator for learning the auxiliary tasks. The behavior policy is set to the target policy of the auxiliary tasks. 
In the second phase, the representation learned during the pre-training phase is kept fixed and fed to another network to learn the main task. The behavior policy in this phase is set to the policy learned for the main task. See Figure \ref{fig:phases}.
Using this procedure, the representation learned for the auxiliary tasks in Phase $1$ is transferred to be used in Phase $2$ when learning the main task.


For the auxiliary tasks, we considered $4$ policies: 1) Uniform random 2) Greedy 3) Sticky actions 4) The main task policy. 
The sticky actions policy is to follow the previous action with probability $0.9$ and to select an action randomly with probability $0.1$.
Learning the auxiliary task corresponding to the greedy policy is a control problem whereas learning the auxiliary tasks corresponding to the other policies are prediction problems. 

We experimented with pixel-based and non pixel-based environments. 
For the non pixel-based environments, we considered two types of cumulants: 1) The observation $O_{t+1}$ 2) The observation difference $O_{t+1} - O_t$. 
We also used $6$ different values of $\gamma$: $\{0, 0.5, 0.75, 0.8, 0.9, 0.99\}$ to cover a wide range of temporal horizons for the auxiliary tasks. 
Therefore, for each policy and each type of cumulant, $6 \times |\mathcal{O}|$ auxiliary tasks were learned in parallel where $|\mathcal{O}|$ denotes the observation size.
To learn the auxiliary tasks in parallel during the pre-training phase, we used a multi-headed network with the representation learning layer being shared between the auxiliary tasks and each head corresponding to one auxiliary task. (In the case of control auxiliary tasks, multiple heads are assigned to each auxiliary task, each corresponding to one state-action value.). 
In the case of prediction auxiliary tasks, the behavior policy during the pre-training phase was set to the target policy of the auxiliary tasks; therefore, learning was on-policy. In the case of control auxiliary tasks, $6 \times |\mathcal{O}|$ target policies were being learned each maximizing the return corresponding to one cumulant and $\gamma$. The behavior policy, in that case, was round-robin over the $6 \times |\mathcal{O}|$ target policies, and learning was off-policy. 

For the pixel-based environments, to specify the auxiliary tasks, we cropped the $40\times40$ observation space into a $24\times24$ region and subdivided it into a $6\times6$ grid of non-overlapping cells of size $4\times4$. 
We considered two types of cumulants: 1) Sum of the pixels in each cell 2) The absolute difference of the sum of the pixels in each cell from $t+1$ to $t$.
We used one value of $\gamma$ equal to $0.9$. 
Therefore, for each policy, $36$ auxiliary tasks were learned in parallel for the pixel-based environments. 
Similar to the non pixel-based environments, for the prediction auxiliary tasks, the behavior policy during pre-training was set to the target policy of the $36$ auxiliary tasks. For the control auxiliary tasks, on the other hand, the behavior policy was round-robin over the $36$ target policies each corresponding to one of the auxiliary tasks. 

\section{Experimental setup and results}
We experimented with classic control environments and pixel-based environments. 
For the classic control environments, we used the Mountain Car \citep{moore1991knowledge} and Acrobot \citep{sutton1995generalization}
problems. 
In the Mountain Car problem, an underpowered car should reach the top of a hill starting from the bottom of the hill. 
The observation space includes the position and velocity of the car with the position in $[-1.2, 0.6]$ and velocity in $[-0.07, 0.07]$. 
The action space includes three actions: 1) throttle forward 2) throttle backward 3) no throttle. The reward is $-1$ in all time steps and $\gamma = 1$. 

Acrobot simulates a two-link underactuated robot with the goal of swinging the endpoint above the bar. 
The observation space includes two angles: 1) the angle between the first link and the vertical line 2) the angle between the two links. 
Note that in the original Acrobot environment, the observation space also includes the angular velocities corresponding to the two angles. 
In our experiments, however, to make the problem more challenging, we omitted the angular velocities. 
This makes the problem partially observable. The action space includes three actions: positive torque, negative torque, and no torque applied to the bottom joint. The reward is $-1$ at all steps and $\gamma = 1$. 

For the pixel-based environments, we used two instances of the Minimalistic Gridworld Environment \citep{gym_minigrid}: 1) Empty Minigrid 2) Door \& Key Minigrid. The objective is to reach the goal cell. 
The observation space is $40\times40\times3$ pixel input. 
The action space includes turning left, turning right, moving forward, picking an object, dropping something off, and opening the door. The reward is $0$ except once the goal is reached in which case it is $1$. $\gamma$ is $0.99$. 

In the pre-training phase, for the classic control environments, we used a neural network with two feed-forward hidden layers of size $100$ with ReLU nonlinearity. 
For Mountain Car, the position and velocity were fed into the network. 
For Acrobot, to deal with partial observability, we passed the angles and the action at the previous time step in addition to the angles at the current time step to the network as input.
For Empty Minigrid, we used a  neural network with one convolutional hidden layer and one feedforward hidden layer of size $64$ with ReLU nonlinearity.
For Door \& Key Minigrid, we used a neural network with three convolutional hidden layers and one feedforward hidden layer of size $32$ with ReLU nonlinearity.
For the phase of learning the main task, we fed the representation learned during pre-training to a network with two feed-forward hidden layers of size $100$.  

We used learning curves to study the effect of different auxiliary tasks on performance. 
We first trained the representation using each of the auxiliary tasks for $500$ episodes for Mountain Car,  Acrobot, and Empty Minigrid, and $2000$ episodes for Door \& Key Minigrid.
To create learning curves, we ran Q-learning with each of the learned representations $30$ times ($30$ independent runs). To get the learning curve for Mountain Car, Acrobot, Empty Minigrid, and Door \& Key Minigrid, we used $400$, $200$, $200$, and $5000$ episodes respectively. 

\begin{figure}[h]
\centering
  \includegraphics[width=1\linewidth]{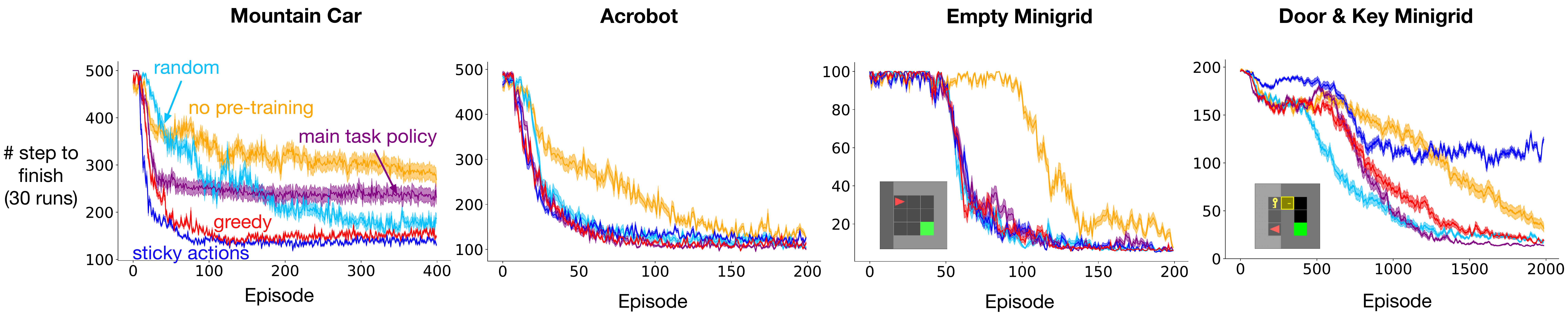}
  \caption{Learning curves for the case of cumulant equal to the observation. In the pre-training phase, the neural network was trained using the auxiliary tasks. The last hidden layer of the neural network was then fixed and used as the representation to learn the main task for $30$ independent runs. In almost all cases, the representation learned for the auxiliary tasks improved over the baseline of no pre-training.
  In Mountain Car, the sticky actions policy and greedy policy resulted in the largest performance gain over the baseline.
  In Acrobot and Empty Minigrid, all auxiliary tasks resulted in similar performance.
  In Door \& Key Minigrid, the random policy resulted in the largest performance gain.}
  \label{fig:obs}
\end{figure}

In the case where the cumulant was equal to the observation, the representation learned for the auxiliary tasks resulted in performance gain over the baseline of no pre-training in almost all cases (Figure \ref{fig:obs}).
In Mountain Car, the sticky actions policy and greedy policy resulted in the largest improvement over the baseline.
In Acrobot and Empty Minigrid, the choice of the target policy did not have a substantial effect.
In Door \& Key Minigrid, the random policy resulted in the largest improvement over the baseline. 

\begin{figure}[h]
\centering
  \includegraphics[width=1\linewidth]{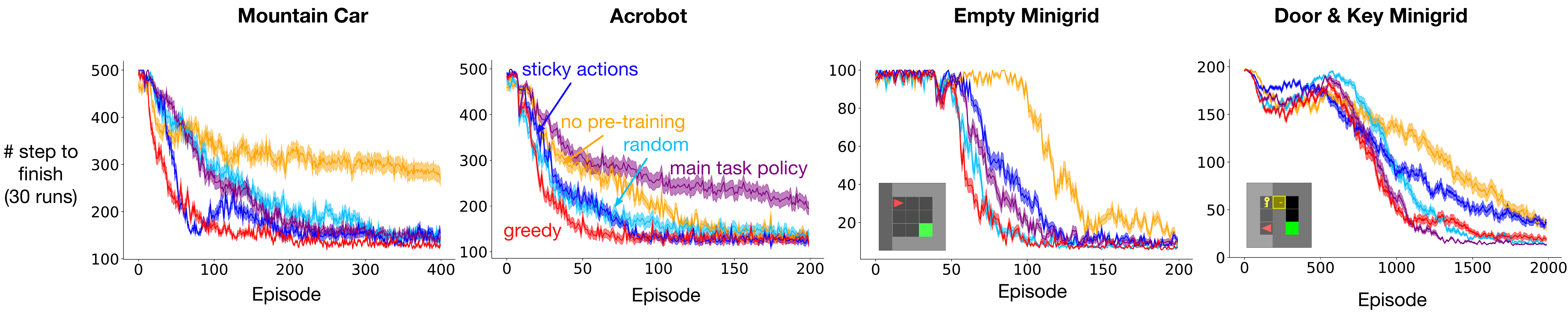}
  \caption{Learning curves for the case of cumulant equal to the observation difference. 
  The results are for $30$ independent runs. 
  In almost all cases, the representation learned for the auxiliary tasks accelerated learning on the main task with the greedy policy resulting in the largest improvement in almost all cases.}
  \label{fig:obs_diff}
\end{figure}

In the case where the cumulant was equal to the observation difference, in almost all cases, the representation learned for the auxiliary tasks resulted in performance gain over the baseline of no pre-training, except for the case where the policy was the main task policy in Acrobot  (Figure \ref{fig:obs_diff}). 
The auxiliary task corresponding to the greedy policy resulted in the highest improvement across tasks.
The results on the other policies were mixed: In the classic control environments, the sticky actions policy resulted in good performance gain whereas the main task policy was less useful or even harmful. 
On the other hand, in the pixel-based environments, the main task policy resulted in good performance whereas the sticky actions policy resulted in worse performance compared to the other auxiliary tasks. 

The results from both classic control and pixel-based environments suggest that the auxiliary tasks based on the greedy policy tend to be useful. 
Note that in the case of pixel-based environments where the cumulant is the observation difference, these auxiliary tasks are similar to the well-known pixel-control auxiliary task.
Another conclusion is that most policies, even the random policy result in performance gain over the baseline. 
A surprising observation is that in most cases, the auxiliary tasks corresponding to the main task policy tend to fall into the group of less useful auxiliary tasks and even in some cases result in worse performance compared to the baseline. 

\section{Conclusions}
Auxiliary tasks about different aspects of the environment can assist representation learning in reinforcement learning. 
The theories about what makes useful auxiliary tasks are still evolving. 
In this paper, we took a step toward answering this question by conducting an empirical study, investigating the role of the target policy in the utility of auxiliary tasks. 
Our study suggests that most policies, including a uniformly random policy, tend to improve over the baseline with the greedy policy resulting in the largest performance gain. 

Further work is required to develop an understanding of how the target policy contributes to the usefulness of the auxiliary tasks. 
A natural future direction is to form concrete hypotheses regarding the observations that we had from our empirical results and answer questions such as:
Why does the greedy policy tend to be useful? 
Why in some cases, the random policy results in the highest performance gain?
Why in many cases, the main task policy results in lower performance gain?

\bibliography{main}

\end{document}